\documentclass[11pt]{article}

\newcommand{\emobank}{\textsc{EmoBank}}
\newcommand{\semeval}{{\sc Se07}}
\newcommand{\subscript}[1]{$_{\text{#1}}$}
\newcommand{\writer}{{\sc Writer}}
\newcommand{\reader}{{\sc Reader}}
\newcommand{\masc}{{\sc Masc}}

\usepackage{EACL-2017-Template/eacl2017}
\usepackage{times}
\usepackage{url}
\usepackage{latexsym}

\usepackage{color}
\usepackage{graphicx}
\usepackage{amsmath}
\usepackage{multirow}

\eaclfinalcopy % Uncomment this line for the final submission
\title{\textsc{EmoBank}: Studying the Impact of Annotation Perspective and Representation Format on Dimensional Emotion Analysis
}

\author{Sven Buechel \and Udo Hahn\\
Jena University Language \& Information Engineering (JULIE) Lab\\
Friedrich-Schiller-Universit\"at Jena, Jena, Germany\\
\url{{sven.buechel,udo.hahn}@uni-jena.de}\\
\url{http://www.julielab.de}
}
\date{}

\begin{document}
\maketitle
\begin{abstract}
We describe \textsc{EmoBank}, a corpus of 10k English sentences balancing multiple genres, which we annotated with dimensional emotion metadata
in the Valence-Arousal-Dominance (VAD) representation format.\ \textsc{EmoBank} excels with a bi-perspectival and bi-representational design.\ On the one hand, we distinguish between writer's and reader's emotions, on the other hand, a subset of the corpus complements dimensional VAD annotations with categorical ones based on Basic Emotions.\ We find evidence for the supremacy of the reader's perspective in terms of IAA and rating intensity, and achieve close-to-human performance when mapping between dimensional and categorical formats.
\end{abstract}

\section{Introduction}
\label{sec:intro}

In the past years, the analysis of affective language has become one of the most productive and vivid areas in computational linguistics. In the early days, the prediction of the semantic polarity (positiveness or negativeness) was in the center of interest, but in the meantime, research activities shifted towards a more fine-grained modeling of sentiment. This includes the extension from only two to multiple polarity classes or even real-valued scores \cite{Strapparava07}, the aggregation of multiple aspects of an opinion item into a composite opinion statement for the whole item \cite{Schouten16}, and sentiment compositionality \cite{Socher13}.

Yet, two important features of fine-grained modeling still lack appropriate resources, namely shifting towards psychologically more adequate models of emotion \cite{Strapparava16} and distinguishing between writer's \textit{vs.}\ reader's perspective on emotion ascription \cite{Calvo13}.
We close both gaps with \emobank{}, the first large-scale text corpus which builds on the Valence-Arousal-Dominance model of emotion, an approach that has only recently gained increasing popularity within sentiment analysis. \textsc{EmoBank} not only excels with a genre-balanced selection of sentences, but is based on a \textit{bi-perspectival} annotation strategy (distinguishing the emotions of writers and readers), and includes a \textit{bi-representationally} annotated subset (which has previously been annotated with Ekman's Basic Emotions) so that mappings between both representation formats can be performed. \emobank{} is freely available for academic purposes.\footnote{\url{https://github.com/JULIELab/EmoBank}}

\section{Related Work}
\label{sec:related}

\begin{figure}
\center
\includegraphics[width = 0.4\textwidth]{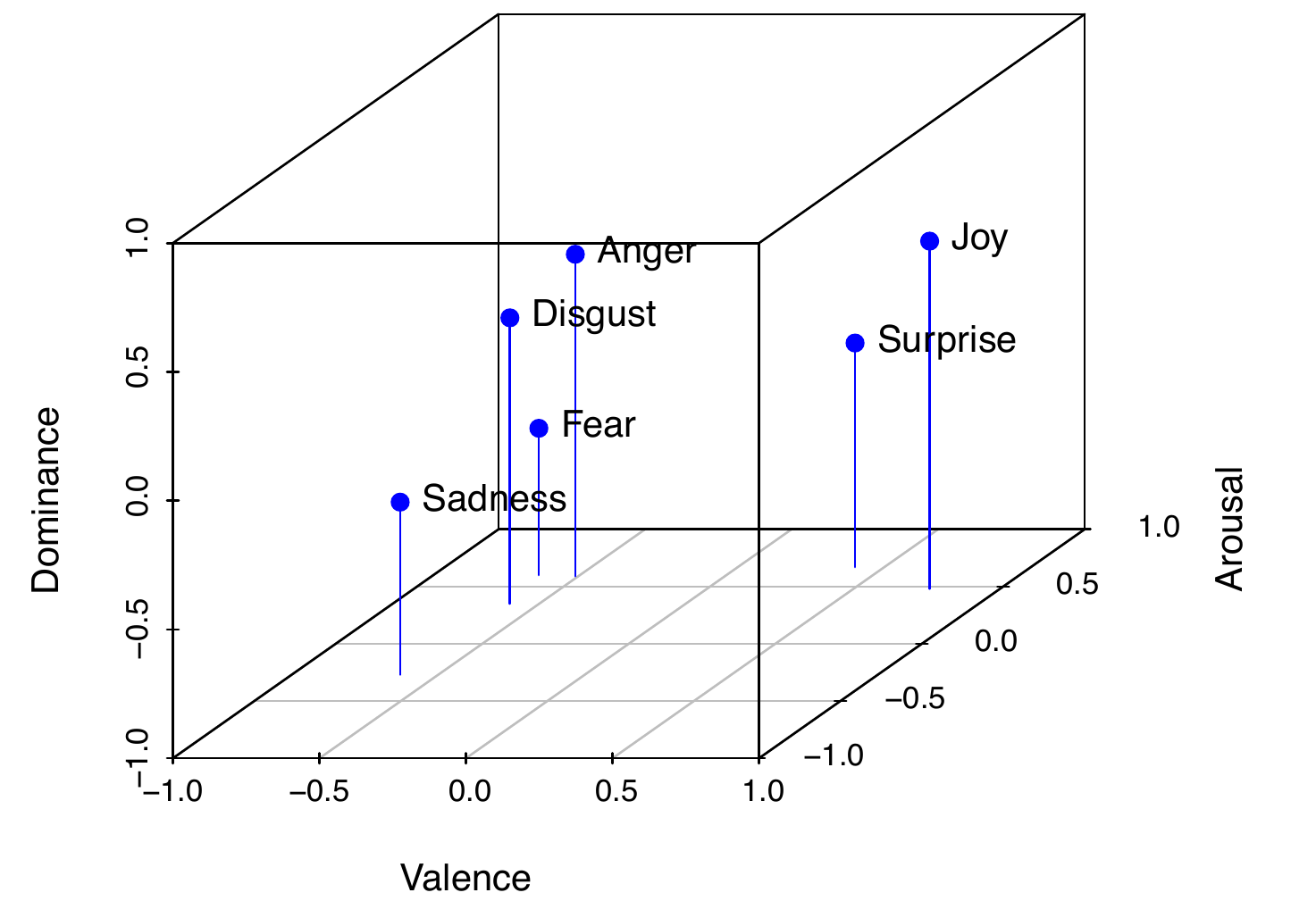}
\vspace*{-5pt}
\caption{
\label{fig:vad}
The affective space spanned by the three VAD dimensions. As an example, we here include the positions of Ekman's six Basic Emotions as determined by \newcite{Russell77}.
}
\vspace*{-10pt}
\end{figure}

Models of emotion are commonly subdivided into {\it categorical} and {\it dimensional} ones, both in psychology and natural language processing (NLP). Dimensional models consider affective states to be best described relative to a small number of independent emotional dimensions (often two or three):  {\it Valence} (corresponding to the concept of polarity), {\it Arousal} (degree of calmness or excitement), and {\it Dominance}\footnote{This dimension is sometimes omitted (the VA model).} (perceived degree of control over a situation); the VAD model. Formally, the VAD dimensions span a three-dimensional real-valued vector space as illustrated in Figure \ref{fig:vad}.
Alternatively, categorical models, such as the six \textit{Basic Emotions} by \newcite{Ekman92} or the \textit{Wheel of Emotion} by \newcite{Plutchik80}, conceptualize emotions as discrete states.\footnote{Both dimensional and categorical formats allow for numerical scores regarding their dimensions/categories.
} 

In contrast to categorical models which were used early on in NLP \cite{Alm05,Strapparava07}, dimensional models have only recently received increased attention in tasks such as word and document emotion prediction (see, e.g., \newcite{Yu15}, \newcite{Koeper16}, %\newcite{Palogiannidi16}, \newcite{Malandrakis13}, \newcite{Shaikh16},
  \newcite{Wang16}, \newcite{Buechel16ecai}).

In spite of this shift in modeling focus, VA(D)-annotated corpora are surprisingly rare in number and small in size, and also tend to be restricted in reliability. \textsc{Anet}, for instance, comprises only 120 sentences designed for psychological research \cite{Bradley07}, while
\newcite{Preotiuc16} created a corpus of 2,895 English Facebook posts relying on only two annotators. \newcite{Yu16} recently presented a corpus of 2,009 Chinese sentences from various online texts. 

As far as categorical models for emotion analysis are concerned, many studies use incompatible subsets of category systems, which limits their comparability \cite{Buechel16ecai,Calvo13}. This also reflects the situation in psychology where there is still no consensus on a set of fundamental emotions \cite{Sander09}. Here, the VAD model has a major advantage: Since the dimensions are designed as being independent, results remain comparable dimension-wise even in the absence of others (e.g., Dominance).
Furthermore, dimensional models are the predominant format for lexical affective resources in behavioral psychology as evident from the huge number of datasets available for a wide range of languages (see, e.g., \newcite{Warriner13}, \newcite{Stadthagen16}, \newcite{Moors13} and \newcite{Schmidtke14}).

%%%%%%%%%%%%%%%%%%%%%%%%%%%%%%%%%%%%%%%%%%%%%%%%
%SB: Die Florida-Gruppe bitte, dass beide Quellen aufgeführt werden.
%%%%%%%%%%%%%%%%%%%%%%%%%%%%%%%%%%%%%%%%%%%%%%%%
For the acquisition of VAD values from participant's self-perception, the Self-Assessment Manikin (SAM; \newcite{Lang80}, \newcite{Bradley94}) has turned out as the most important and (to our knowledge) only standardized instrument \cite{Sander09}. SAM iconically displays differences in Valence, Arousal and Dominance by a set of anthropomorphic cartoons on a multi-point scale (see Figure \ref{fig:SAM}).

While it is common for more basic sentiment analysis systems in NLP to map the many different possible interpretations of a sentence's affective meaning into a single assessment (``its sentiment''), there is an increasing interest in a more fine-grained approach where emotion expressed by writers is modeled separately from emotion evoked in readers. An utterance like ``Italy defeats France in the World Cup Final'' may be completely neutral from the \textit{writer's} viewpoint (presumably a professional journalist), but is likely to evoke rather adverse emotions in Italian and French \textit{readers} \cite{Katz07}. 

In this line of work, \newcite{Tang12} examine the relation between the sentiment of microblog posts and the sentiment of their comments (as a proxy for reader emotion).
\newcite{Liu13} model the emotion of a news reader jointly with the emotion of a comment writer using a co-training approach. This contribution was followed up by \newcite{Li16} who propose a two-view label propagation approach instead. 
However, to our knowledge, only \newcite{Mohammad13} investigated the effects of these perspectives on annotation quality, finding differences in inter-annotator agreement (IAA) relative to the exact phrasing of the annotation task.

In a similar vein to the writer-reader distinction, identifying the {\it holder} or {\it source} of an opinion or sentiment also aims at describing the affective information entailed in a sentence in more detail \cite{Wiebe05,Seki09}. Thus, opinion statements that can directly be attributed to the writer can be distinguished from references to other's opinions.
A related task, the detection of {\it stance}, focuses on inferring the writer's (dis)approval  towards a given issue from a piece of text \cite{Sobhani16}.

\section{Corpus Design and Creation}
\label{sec:acquisition}

The following criteria guided the data selection process of the \textsc{EmoBank} corpus:
First, complementing existing resources which focus on social media and/or review-style language \cite{Yu16,Quan09}, we decided to address several genres and domains of general English.

Second, we conducted a pilot study on two samples (one consisting of movie reviews, the other pulled from a genre-balanced corpus) to compare the IAA resulting from different annotation perspectives 
(e.g., the writer's and the reader's perspective) in different domains (see \newcite{Buechel17law} for details). Since we found differences in IAA but the results remained inconclusive, we decided to annotate the whole corpus {\it bi-perspectively}, i.e., each sentence was rated according to both the (perceived) writer {\it and} reader emotion (henceforth, \writer{} and \reader{}).

Third, since many problems of comparing emotion analysis studies result from the diversity of emotion representation schemes (see Section \ref{sec:related}), the ability to accurately map between such alternatives would greatly improve comparability across systems and boost the re-usability of resources. Therefore, at least parts of our corpus should be annotated \textit{bi-representationally} as well, complementing dimensional VAD ratings with annotations according to a categorical emotion model.

Following these criteria, we composed our corpus out of several categories of the \textit{ Manually Annotated Sub-Corpus  of the American National Corpus} (\masc{}; \newcite{Ide08}, \newcite{Ide10}) and the corpus of SemEval-2007 Task 14 \textit{Affective Text} (\semeval{}; \newcite{Strapparava07}). \masc{} is already annotated on various linguistic levels. Hence, our work will allow for research at the intersection of emotion and other language phenomena. \semeval{}, on the other hand, bears annotations according to Ekman's six Basic Emotion (see Section \ref{sec:related}) on a $[0, 100]$ scale, respectively. This collection of raw data comprises 10,548 sentences (see  Table \ref{tab:composition}).

\begin{table}[]
\center
{\small
\begin{tabular}{|l|l|r|r|}
\hline
Corpus & Domain & Raw & Filtered \\
\hline
\hline
\semeval{} &	news headlines & 1,250	&	1,192\\
\hline
\multirow{6}{*}{\masc{}} & blogs	&1,378&	1,336\\
&essays &1,196	&	1,135\\
&fiction &2,893	&	2,753\\
&letters	&1,479	&	1,413\\
&newspapers	&1,381	&	1,314\\
&travel guides	&971	&	919\\
\hline
\hline
\multicolumn{2}{|l|}{\textbf{Sum}} & \textbf{10,548} &	\textbf{10,062}\\
\hline	
\end{tabular}
}
\caption{\label{tab:composition} Genre distribution of the raw and filtered \emobank{} corpus.}\vspace{-5pt}
\end{table}

Given this large volume of data, we opted for a crowdsourcing approach to annotation. We chose \textsc{CrowdFlower} (\textsc{Cf}) over \textsc{Amazon Mechanical Turk} (\textsc{Amt}) 
for its quality control mechanisms and accessibility (customers of \textsc{Amt}, but not \textsc{Cf}, must be US-based).\
\textsc{Cf}'s main quality control mechanism rests on \textit{gold questions}, items for which the acceptable ratings have been previously determined by the customer. These questions are inserted into a task to restrict the workers to those performing trustworthily. 
We chose these gold items by automatically extracting highly emotional sentences from our raw data according to {\sc JEmAS}\footnote{\url{https://github.com/JULIELab/JEmAS}}, a lexicon-based tool for VAD prediction \cite{Buechel16ecai}. The acceptable ratings were determined based on manual annotations by three students trained in linguistics. The process was individually performed for \writer~and \reader~with different annotators.

\begin{figure}[th]
\center
\includegraphics[width=.4\textwidth]{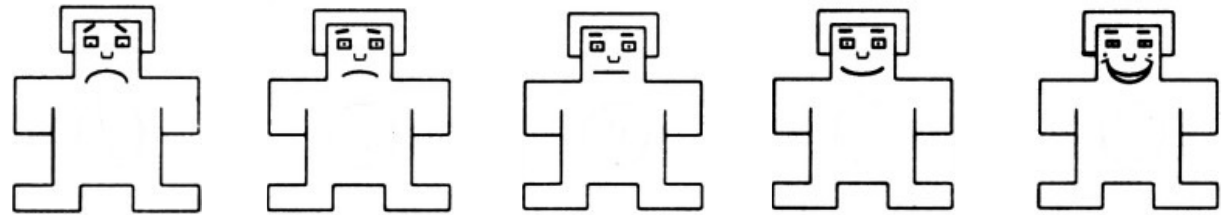}
\vspace*{-0pt}

\hspace*{2pt}\includegraphics[width=.4\textwidth]{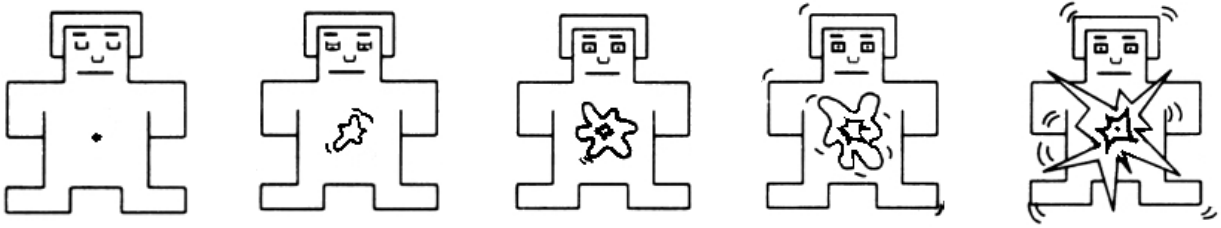}
\vspace*{-0pt}

\hspace*{3pt}\includegraphics[width=.4\textwidth]{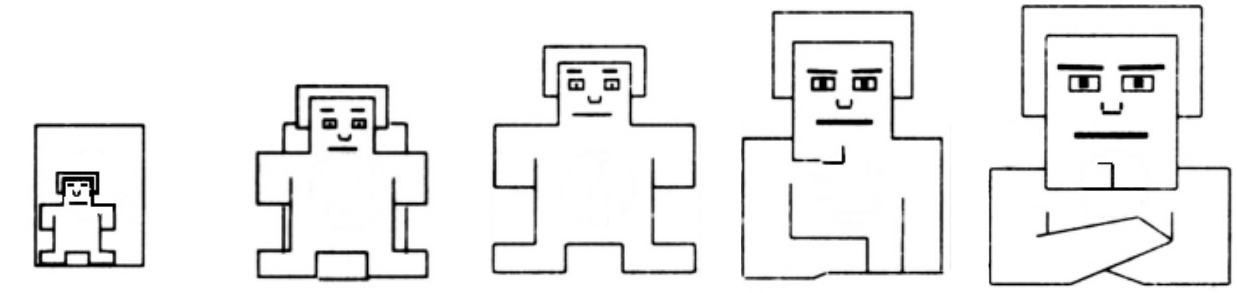}\vspace{-5pt}

\caption{\label{fig:SAM} The modified 5-point Self-Assessment Manikin (SAM) scales for Valence, Arousal and Dominance (row-wise). Copyright of the original SAM by Peter J.\ Lang 1994.}
\vspace*{-9pt}
\end{figure}

For each of the two perspectives, we launched an independent task on \textsc{Cf}. The instructions (see Appendix) were based on those by \newcite{Bradley99anew} to whom most of the VAD resources developed in psychology refer (see Section \ref{sec:related}). We changed the 9-point SAM scales to 5-point scales (see Figure \ref{fig:SAM}) in order to reduce the cognitive load during decision making for crowdworkers. For the writer's perspective, we presented a number of linguistic clues supporting the annotators in their rating decisions, while, for the reader's perspective, we asked what emotion would be evoked in an {\it average} reader (rather than asking for the rater's personal feelings).
Both adjustments were made to establish more objective criteria for the exclusion of untrustworthy workers. We provide the instructions along with our dataset.

For each sentence, five annotators generated VAD ratings. Thus, a total of 30 ratings were gathered per sentence (five ratings for each of the three VAD dimensions and two annotation perspectives, \writer{} and \reader{}). Ten sentences were presented at a time. The task was available for workers located in the UK, the US, Ireland, Canada, Australia or New Zealand. The total annotation costs amounted to \$1,578.

Upon inspection of the individual judgments, we found that the VAD rating $(1,1,1)$ was heavily overrepresented. 
We interpret this skewed coding distribution as a bias mainly due fraudulent responses since, from a psychological view, this rating is highly improbable \cite{Warriner13}.
Accordingly, we decided to remove all of these ratings (about 10\% for each of the tasks; the `Filtered' condition in Table \ref{tab:composition}) because these annotations would have inserted a systematic bias into our data which we consider more harmful than erroneously removing a few honest outliers.
For each sentence with two or more remaining judgments, its final emotion annotation is determined by averaging these valid ratings leading to a total of 10,062 sentences bearing VAD values for \textit{both} perspectives (see Table \ref{tab:composition}). 

This makes \emobank{} to the best of our knowledge by far the largest corpus for dimensional emotion models and, with the exception of the dataset by \newcite{Quan09} (which is problematic in having only {\it one} annotator per sentence), the largest gold standard for any emotion format (both dimensional and categorical). Even compared with polarity corpora it is still reasonably large (e.g., similar in size to the \textit{Stanford Sentiment Treebank} \cite{Socher13}).

\section{Analysis and Results}
\label{sec:results}

For continuous, real-valued numbers, well-known metrics for IAA, such as Cohen's $\kappa$ or F-score, are inappropriate as these are designed for nominally scaled variables. Instead, Pearson's correlation coefficient ($r$) or Mean Absolute Error (\textit{MAE}) are often applied for this setting \cite{Strapparava07,Yu16}. Accordingly, for each annotator, we compute $r$ and \textit{MAE} between their own and the aggregated \emobank{} annotation and average these values for each VAD dimension. This results in one IAA value per metric ($r$ or {\it MAE}), perspective and dimension (Table \ref{tab:iaa}).

As average over the VAD dimensions, we achieve a satisfying IAA of $r>.6$ for both perspectives. The \reader{} results in significantly higher correlation,\footnote{Note that using this set-up, obtaining statistical significance is very rare, since the number of cases is based on the number of raters.} but also higher error than \writer{} ($ p < .05$ for Valence in $r$ and for all dimensions in \textit{MAE} using a two-tailed $t$-test).

\begin{table}[]
\center
\small
\begin{tabular}{|c||c|c|c||c|}
\hline
	&	Valence 	&	Arousal 	& 	Dominance 	&	\textbf{Av.}\\
\hline
\hline
$r_{\text{writer}}$	& 	0.698	&	0.578	&	0.540	&	0.605\\
$r_{\text{reader}}$	&	0.738	&	0.595	&	0.570	&	0.634\\
\hline
$\text{\textit{MAE}}_{\text{writer}}$&0.300&0.388&0.316&0.335\\
$\text{\textit{MAE}}_{\text{reader}}$&0.349&0.441&0.367&0.386\\
\hline
\end{tabular}
\caption{IAA for the three VAD dimensions.\label{tab:iaa}}
%\vspace{-10pt}
\end{table}

Prior work found that a large portion of language may actually be neutral in terms of emotion \cite{Alm05}. However, a too narrow rating distribution (i.e., most of the ratings being rather neutral relative to the three VAD dimensions) may be a disadvantageous property for training data. Therefore, we regard the {\it emotionality} of ratings as another quality criterion for emotion annotation complementary to IAA.  

We capture this notion as the absolute difference of a sentence's aggregated rating from the neutral rating ($3$, in our case), averaged over all VAD dimensions. Comparing the average  emotionality of all sentences between \writer{} and \reader{}, we find that the latter perspective also excels with significantly higher emotionality than the \writer{} ($p < .001$; two-tailed $t$-test). 

These beneficial characteristics of the \reader{} perspective (better correlation-based IAA and emotionality) contrast with its worse error-based IAA. Thus, we decided to examine the relationship between error and emotionality between the two perspectives more closely: 
Let $V,A,D$ be three $m\times n$-matrices where $m$ corresponds to the number of sentences and $n$ to the number of annotators so that the three matrices yield all the individual ratings for Valence, Arousal and Dominance, respectively. Then we define the {\it sentence-wise error} for sentence $i$ ($\text{SWE}_i$) as 
\begin{equation}
\text{SWE}_i := \frac{1}{3}\sum_{X \in \{V,A,D\}} \frac{1}{n}\sum_{j=1}^n |\overline{X_i} - X_{ij}|
\end{equation}
where $\overline{X_i} := \frac{1}{n}\sum_{j=1}^n X_{ij}$.
We compute SWE values for reader and writer perspective individually. We can now examine the dependency between error and emotionality by subtracting, for each sentence, SWE and emotionality for both perspectives from another  (resulting in one {\it difference in error} and one {\it difference in emotionality} value).

\begin{figure}
\center
\includegraphics[width = .35\textwidth]{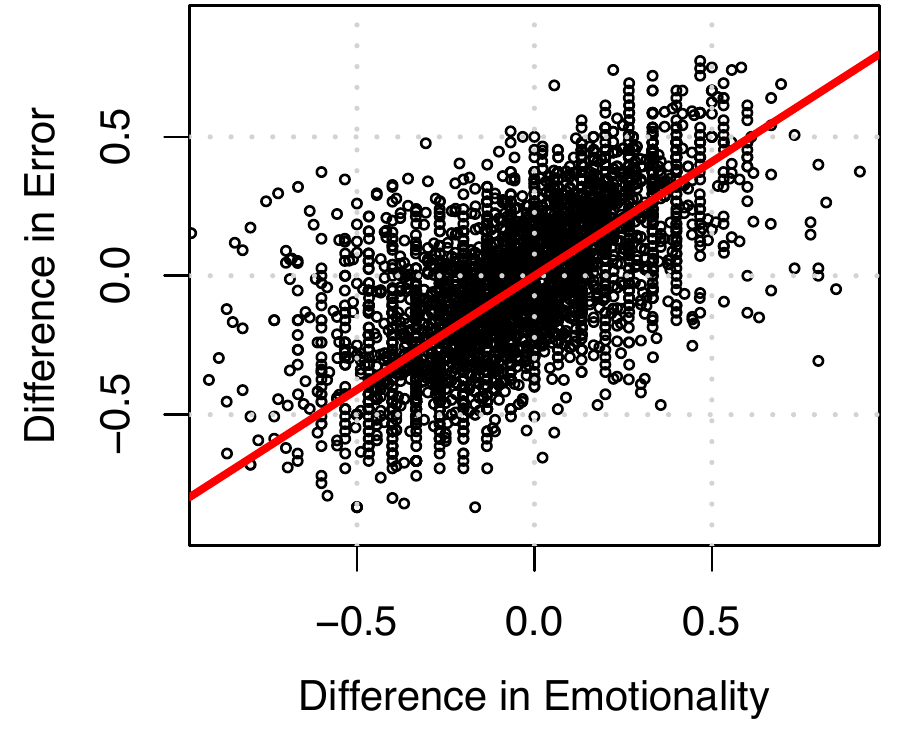}
\vspace{-10pt}
\caption{Differences in emotionality and differences in error between \writer{} and \reader{}, each sentence corresponding to one data point; regression line depicted in red.\label{fig:lm}}
\vspace*{-10pt}
\end{figure}

Our data reveal a strong correlation ($r = .718$) between these data series, so that the more the ratings for a sentence differ in emotionality (comparing between the perspectives), the more they differ in error as well. Running linear regression on these two data rows, we find that the regression line runs straight through the origin (intercept is \textit{not} significantly different from 0; $p=.992$; see Figure \ref{fig:lm}). This means that without difference in emotionality, \writer{} and \reader{} rating for a sentence do, on average, {\it not} differ in error. Hence, our data strongly suggest that \reader{} is the superior perspective yielding better inter-annotator {\it correlation} and emotionality without overproportionally increasing inter-annotator {\it error}.

\section{Mapping between Emotion Formats}

Making use of the bi-representational subset of our corpus (\semeval{}), we now examine the feasibility of automatically mapping between dimensional and categorical models. For each Basic Emotion category, we train one $k$ Nearest Neighbor model given all VAD values of either \writer{}, \reader{} or both combined as features. Training and hyper-parameter selection was performed using 10-fold cross-validation. 

Comparing the correlation between our models' predictions and the actual annotations (in categorical format) with the IAA as reported by \newcite{Strapparava07}, we find that this approach already comes close to human performance (see Table \ref{tab:mapping}). 
Once again, \reader{} turns out to be superior in terms of the achieved mapping performance compared to \writer{}. However, both perspectives combined yield even better results. 
In this case, our models' correlation with the actual \semeval{} rating is as good as or even better than the average human agreement. Note that the \semeval{} ratings are in turn based on averaged human judgments.  
Also, the human IAA differs a lot between the Basic Emotions and is even $r < .5$ for Disgust and Surprise. For the four categories  with a reasonable IAA, Joy, Anger, Sadness and Fear, our best models, on average, actually outperform  human agreement. Thus, our data shows that automatically mapping between representation formats is feasible at a performance level on par with or even surpassing human annotation capability. This finding suggests that, for a dataset with high-quality annotations for one emotion format, automatic mappings to another format may be just as good as creating these new annotations by manual rating. 

\begin{table}[t]
\small
\center
\begin{tabular}{|l||c|c|c|c|c|c||c|c|}
  \hline
 & Joy & Ang & Sad & Fea & Dsg & Srp & {\bf Av.} \\ 
  \hline \hline
\hspace{-3pt}IAA\hspace{-10pt} & .60 & .50 & .68 & .64 & .45 & .36 & .54\hspace{-3pt} \\ 
\hline
  \hspace{-3pt}W\hspace{-10pt} & .68 & .40 & .67 & .47 & .27 & .15 & .44\hspace{-3pt} \\ 
  \hspace{-3pt}R\hspace{-10pt} & .73 & .47 & .68 & .54 & .36 & .15 & .49\hspace{-3pt} \\ 
 \hspace{-3pt}WR\hspace{-10pt} & .78 & .50 & .74 & .56 & .36 & .17 & .52\hspace{-3pt} \\
  \hline 
  \hspace{-3pt}D\subscript{W}\hspace{-10pt} & +.08 & --.10 & --.01 & --.17 & --.17 & --.21 & --.09\hspace{-3pt}  \\ 
  \hspace{-3pt}D\subscript{R}\hspace{-10pt}& +.13 & --.03 & +.00 & --.10 & --.09 & --.22 & --.05\hspace{-3pt} \\ 
 \hspace{-3pt}D\subscript{WR}\hspace{-10pt} & +.18 & +.00 & +.05 & --.08 & --.09 & --.19 & --.02\hspace{-3pt} \\ 
   \hline
\end{tabular}
\caption{\label{tab:mapping} IAA by \newcite{Strapparava07} compared to mapping performance of KNN models using writer's, reader's or both's VAD scores as features (W, R and WR, respectively), both in Pearson's $r$. Bottom section: difference of respective model performance (W, R and WR) and IAA.}%\vspace*{-7pt}
\end{table}

\section{Conclusion}
\label{sec:conclusion}

We described the creation of \textsc{EmoBank}, the first large-scale corpus employing the dimensional VAD model of emotion and one of the largest gold standards for {\it any} emotion format.
This genre-balanced corpus is also unique for having two kinds of double annotations. First, we annotated for both  writer and reader emotion; second, for a subset of the \textsc{EmoBank},
ratings for categorical Basic Emotions as well as VAD dimensions are now available.
The statistical analysis of our corpus revealed that the reader perspective yields both better IAA values and more emotional ratings. For the bi-representationally annotated subcorpus, we showed that an automatic mapping between categorical and dimensional formats is feasible with near-human performance using standard machine learning techniques.

\section*{Acknowledgments}%\vspace*{-5pt}
We thank The Center for the Study of Emotion and Attention, University of Florida, for granting us access to the Self-Assessment-Manikin (SAM).

\bibliographystyle{EACL-2017-Template/eacl2017}
\bibliography{literatureSB-EACL2017}

%kein appendix in final
\clearpage
\appendix

\section{Appendix: Instructions} 
\label{app:instructions}

In the following, the instructions for the two \textsc{CrowdFlower} tasks for annotating \textsc{EmoBank} with author emotion (``Expressing Emotion'') and reader emotion (``Evoking Emotion'') are presented. 
These instructions are based on the ones used by \newcite{Bradley99anew} for creating their highly influential VAD lexicon (see Section \ref{sec:related}). Our major modifications are, first, changing from 9-point VAD scales to 5-point scales in order to reduce the cognitive load for the crowd workers and therefore make the task more feasible for crowdsourcing, and, second, changing the instructions towards asking for reader and author perspective, respectively. 

For the author's perspective, we presented a number of linguistic clues (``Hints'') supporting the annotators in their rating decisions, while, for the reader's perspective, we asked for the emotion of an \textit{average} reader instead of their personal feelings. Both adjustments were made to establish more objective criteria for the exclusion of untrustworthy workers (see Section \ref{sec:acquisition}).
Furthermore, we used the alternative names of the Valence-Arousal-Dominance dimensions, Pleasure-Arousal-Control, in order to make their meaning intuitively more clear. 

\subsection*{Expressing Emotion}

\paragraph{Getting Started.} Thank you for participating in this task. Your contribution to academic research is highly appreciated. The study being conducted here is investigating emotion, and specifically concerns how emotion is expressed in different sentences.

\paragraph{Instructions.}
Below, you can see three rows of figures which we call SAM. SAM shows three different kinds of feelings: Unhappy \textit{vs.}\ happy (``Pleasure''), calm \textit{vs.}\ excited (``Arousal'') and submissive \textit{vs.}\ dominant (``Control''). You will be using three independent multiple-choice scales (corresponding to these three panels below) to rate what kind of feeling the author expresses by each sentence (how does he or she feel while writing it?).

\begin{figure}[h!]
\center
\large
Pleasure
\includegraphics[width = 0.45\textwidth]{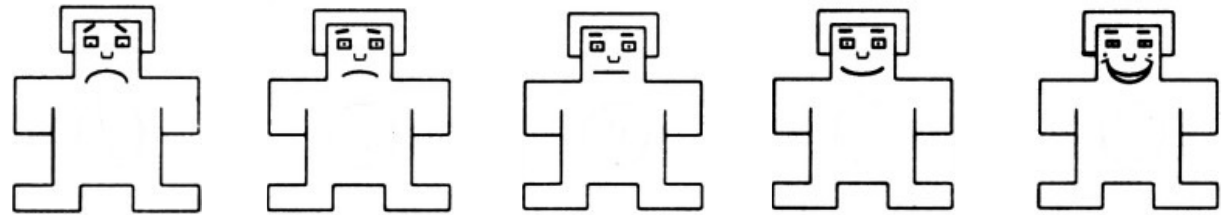}
\end{figure}

The first panel (``Pleasure'') shows an unhappy-happy scale. On the left side of this scale, the author of a sentence expresses unhappiness, annoyance, unsatisfaction, melancholia or boredom. If the author expresses complete unhappiness, please, indicate this by choosing the leftmost option. On the right side of the scale, the author expresses happiness, pleasure, satisfaction, content or hope. If complete happiness is expressed, you can indicate this by choosing the rightmost option. If a sentence is completely neutral, neither expressing happiness nor sadness, please, choose the middle option. You can indicate intermediate levels of happiness or unhappiness by choosing the intermediate options.

Please, take a moment to familiarize yourself with the pictograms of the Pleasure scale.

\begin{figure}[h!]
\center
\large
Arousal
\includegraphics[width = 0.45\textwidth]{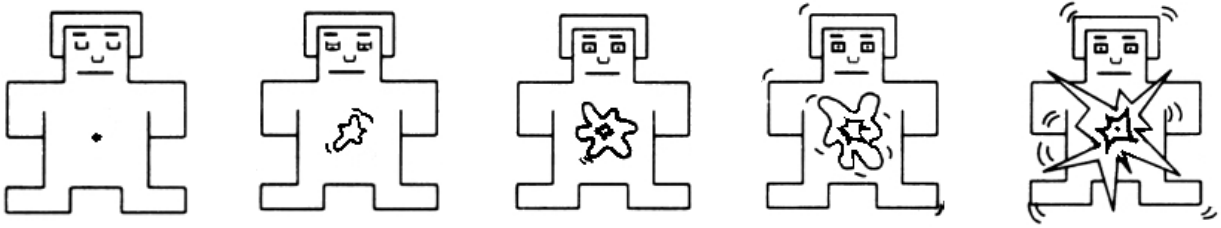}
\end{figure}

The second panel (``Arousal'') shows a calm-excited scale. On the left side of this scale, the author of a sentence expresses complete relaxation, calmness, sluggishness, dullness, sleepiness, or unexcitement. If complete calmness is expressed, please, indicate this by choosing the leftmost option. On the right side of the scale, the author expresses stimulation, excitement, frenzy, nervousness, or arousal. If complete arousal is expressed, please, choose the rightmost option. If neither excitement nor calmness is expressed, please, choose the middle option. Choose in-between options to indicate intermediate levels of excitement or calmness.

Please, take a moment to familiarize yourself with the pictograms of the Arousal scale.

\begin{figure}[h!]
\center
\large
Control
\includegraphics[width = 0.45\textwidth]{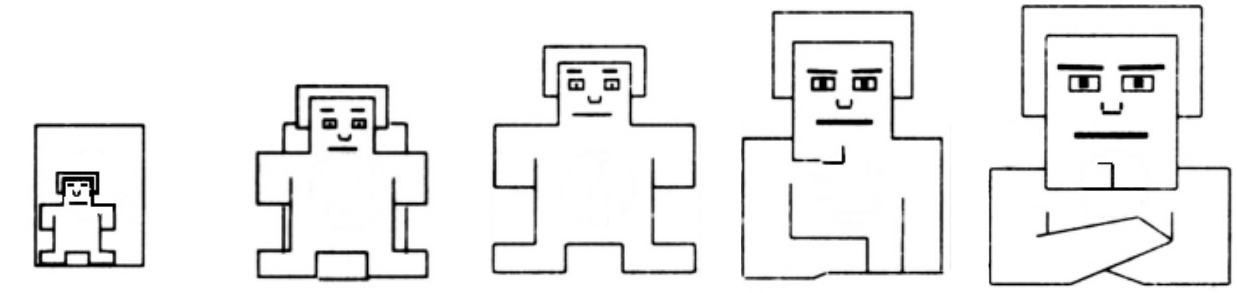}
\end{figure}

The last scale of feeling which you will rate (``Control'') focuses on whether the author expresses submissiveness or dominance. On the left side of the scale, the author has feelings characterized as controlled, influenced, cared-for, awed, submissive, or guided. Please, indicate that the author expresses complete submissiveness by choosing the leftmost option. On the right side of the panel, the author expresses feeling in-control, influential, important, dominant, autonomous, or controlling. If the author feels completely dominant, please, choose the rightmost option. Again, if the author feels neither controlling nor controlled please choose the middle option. You can rate intermediate levels by choosing the options in between.

Please take a moment to familiarize yourself with the pictograms of the Control scale.

\paragraph{Hints.}
The following is meant as rules of thumb to help you with your judgments specifically in the beginning of the task.
\begin{itemize}

	\item Sentences in past tense or 3rd person tend to express rather neutral feelings on the author's side (on contrast to present tense and 1st person) because he or she is less directly involved.

	\item Capitalization, exclamation marks and swearing may often hint at high Arousal.

	\item Giving commands is a reliable indicator for high Control. 
	
	\item Aggressive/strong language or swearing also often translates into high control (as this is linked to dominant behavior).
	
	\item When you think the author feels as always (maybe  just as you feel when, say, walking down the street), that's a 3 on \textit{each of the scales}. Low Arousal values are reserved for feeling specifically relaxed (like in the bathtub), while low Control values are appropriate when you feel specifically powerless (like when being tied up).

\end{itemize}

\paragraph{Final Note.}
Again, please, concentrate fully on what kind of feeling the author expresses with a given sentence rather than how you feel when reading it. If a sentence could be given a high or a low rating on one of the three scales depending on its interpretation, please, decide on the most likely interpretation instead of using the neutral (middle) rating slot by default.

Thank you for reading the instructions carefully. On the next pages, you will be presented sentences which are taken from various sources. Please, rate each of the sentences according to the three SAM scales. Try to use the full range of the scales and work on a rapid pace rather than overinterpreting or spending too much time thinking about the individual sentences.

%++++++++++ Zweite Satz von Instructions ++++++++++%

\subsection*{Evoking Emotion}

\paragraph{Getting started.}
Thank you for participating in this task. Your contribution to academic research is highly appreciated. The study being conducted here is investigating emotion, and is concerned with how people react to different sentences.

\paragraph{Instructions.}
Below, you can see three rows of figures which we call SAM. SAM shows three different kinds of feelings: Unhappy \textit{vs.}\ happy (``Pleasure''), calm \textit{vs.}\ excited (``Arousal'') and submissive \textit{vs.}\ dominant (``Control''). You will be using three independent multiple-choice scales (corresponding to these three panels below) to rate how people feel when reading each sentence. Try to imagine how people would react, on average, while ignoring your personal opinions, believes and experiences as good as you can.

\begin{figure}[h!]
\center
\large
Pleasure
\includegraphics[width = 0.45\textwidth]{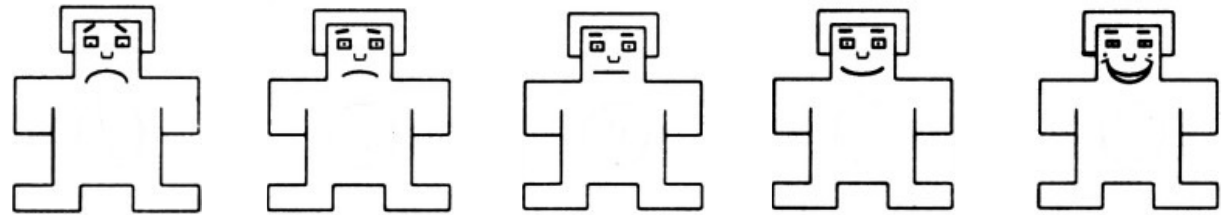}
\end{figure}

The first panel (``Pleasure'') shows an unhappy-happy scale. On the left side of this scale, people feel unhappy, annoyed, unsatisfied, melancholic or bored. If you think people would feel completely unhappy, please, indicate this by choosing the leftmost option. On the right side of the scale, people are happy, pleased satisfied, contented or hopeful. If people would feel completely happy, you can indicate this by choosing the rightmost option. If you think people would feel completely neutral, neither happy nor sad, please, choose the middle option. You can indicate intermediate levels of happiness or unhappiness by choosing the options in between.

Please, take a moment to familiarize yourself with the pictograms of the Pleasure scale.

\begin{figure}[h!]
\center
\large
Arousal
\includegraphics[width = 0.45\textwidth]{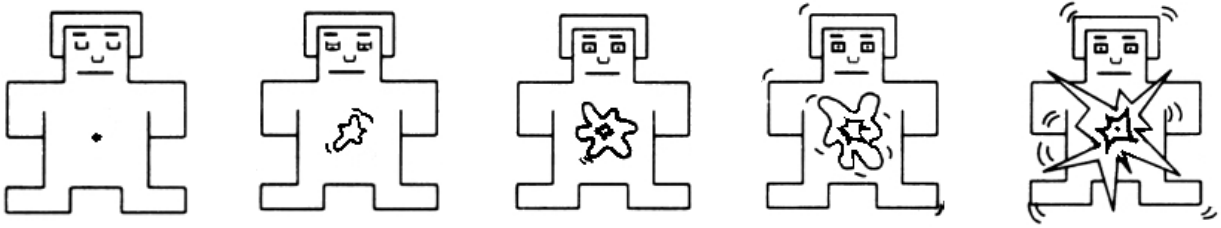}
\end{figure}

The second panel (``Arousal'') shows a calm-excited scale. On the left side of this scale, people feel completely relaxed, calm, sluggish, dull, sleepy, or unaroused. If you think people would feel completely calm, please, indicate this by choosing the leftmost option. On the right side of the scale, people are stimulated, excited, frenzied, jittery, wide-awake, or aroused. When you think people would feel completely aroused, please, choose the rightmost option. If you think people would not be excited nor at all calm, please choose the middle option. Choose in-between options to indicate intermediate levels of excitement or calmness.

Please, take a moment to familiarize yourself with the pictograms of the Arousal scale.

\begin{figure}[h!]
\center
\large
Control
\includegraphics[width = 0.45\textwidth]{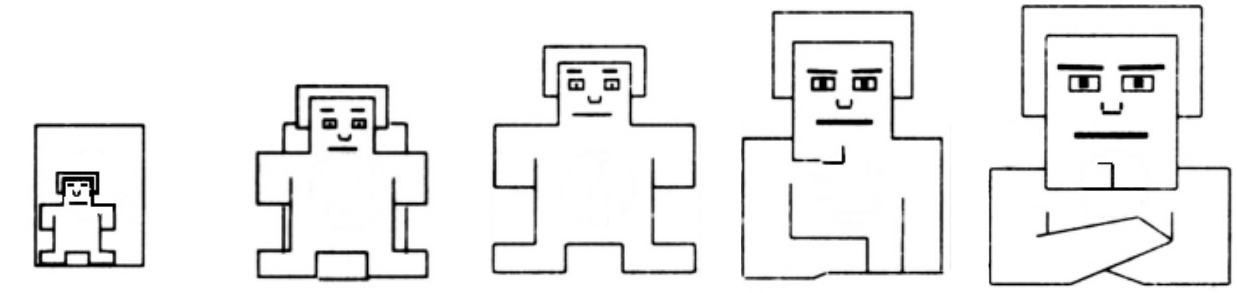}
\end{figure}

The last scale of feeling which you will rate (``Control'') captures whether people feel controlled or in control. On the left side of the scale, people have feelings characterized as controlled, influenced, cared-for, awed, submissive, or guided. Please, indicate that people would feel completely controlled by choosing the leftmost option. On the right side of the panel, people feel in control, influential, important, dominant, autonomous, or controlling. If you think people would feel completely dominant, please, choose the rightmost option. Again, if you think people would feel neither in control nor controlled, please, choose the middle option. Also, you can rate intermediate levels by choosing the options in between.

Please, take a moment to familiarize yourself with the pictograms of the Control scale.

\paragraph{Final Note.}

Again, please, concentrate on how people \textit{in general} would feel after reading a given sentence rather than on your personal reaction or what you think the author felt when writing it. If a sentence could be given a high or a low rating on one of the three scales depending on its interpretation, please, decide for the most likely interpretation instead of using the neutral (middle) rating slot by default.

Thank you for reading the instructions carefully. On the next pages, you will be presented sentences which are taken from various sources. Please, rate each of the sentences according to the three SAM scales. Try to use the full range of the scales and work on a rapid pace rather than overinterpreting or spending too much time thinking about the individual sentences.

\end{document}